\documentclass[a4paper, 10pt, twocolumn]{article}

\usepackage{amsfonts,amssymb,amsmath,amsthm}
\usepackage{subfigure}
\usepackage{graphicx}
\usepackage[footnotesize]{caption}

\usepackage[top=1.5cm, left=1.5cm, right=1.5cm, bottom=1.5cm]{geometry}

\newcommand{\ts}{\textstyle}
\newcommand{\bs}{\boldsymbol}
\newcommand{\cl}{\mathcal}
\newcommand{\bb}{\mathbb}
\newcommand{\wh}{\widehat}
\renewcommand{\Vec}[1]{\bs{#1}} % vector, requires bm package
\newcommand{\expec}[1]{\mathop{{}\mathbb{E}}_{#1}} % for mathematical expectation

\newcommand{\im}{\mathrm{i}\mkern1mu} % Complex number i

\usepackage[style=unsrt,url=false,isbn=false,doi=false,style=numeric-comp,giveninits,backend=biber,sorting=none]{biblatex}
\newcommand{\ie}{\emph{i.e.}, }
\newcommand{\eg}{\emph{e.g.}, }
\newcommand{\iid}{%
    \ifmmode% math mode
        \mathrm{i.i.d.}%
    \else%
        i.i.d.\@\xspace%
    \fi%
}

\usepackage{ifthen}
\newif\ifcomment
\commenttrue
\ifcomment
\newcommand{\LJc}[1]{\textcolor{blue}{\textbf{\scriptsize [LJ: #1]}}}
\newcommand{\VSc}[1]{\textcolor{SeaGreen}{\textbf{\scriptsize [VS: #1]}}}
\newcommand{\todo}[1]{\textcolor{red}{\textbf{\scriptsize [TODO: #1]}}}
\newcommand{\removable}[1]{\textcolor{green}{#1}}
\else
\newcommand{\LJc}[1]{}
\newcommand{\VSc}[1]{}
\newcommand{\todo}[1]{}
\newcommand{\removable}[1]{}
\fi

\title{When compressive learning fails: blame the decoder or the sketch?}

\author{Vincent Schellekens$^1$ and Laurent Jacques$^{1}$.\\
\footnotesize $^1$ICTEAM/ELEN, UCLouvain.
}
\date{\empty} % no need for a date

\renewenvironment{abstract}{\bf\small {\em\ Abstract---}}{}

\begin{document}

\maketitle

\begin{abstract} In compressive learning, a mixture model (a set of centroids or a Gaussian mixture) is learned from a sketch vector, that serves as a highly compressed representation of the dataset. This requires solving a non-convex optimization problem, hence in practice approximate heuristics (such as \texttt{CLOMPR}) are used. In this work we explore, by numerical simulations, properties of this non-convex optimization landscape and those heuristics.
\end{abstract}
% target: 0.25 column

\section{Introduction}
\label{sec:introduction}
% The compressive learning context
The general goal of machine learning is to infer a model, usually represented by parameters $\Vec{\theta}$, from a dataset $X = \{\Vec{x}_i\}_{i =1}^n$ of $n$ training examples $\Vec{x}_i \in \bb R^d$. When $n$ is very large (as in many modern datasets), conventional learning algorithms---that operate in several passes over $X$---require enormous computational resources (both in memory and time). Compressive learning (CL) tackles this critical challenge by first (in a single parallelizable pass) compressing $X$ into a lightweight vector $\Vec{z} \in \bb C^m$ called the \textit{sketch}~\cite{gribonval2017compressiveStatisticalLearning}. The parameters $\wh{\Vec{\theta}} = \Delta[\Vec{z}]$ are then learned by a \textit{decoder} algorithm $\Delta$ from only this compressed proxy for $X$, using potentially much less resources than operating on the raw dataset $X$.

% Sketching in practice and where it could go wrong
In general, the sketch of a distribution $\cl P$ is a linear operator, constructed as a set of generalized moments $\cl A(\cl P) := \expec{\Vec{x} \sim \cl P} \Phi(\Vec{x})$ where the feature map $\Phi$ is typically randomly generated; the sketch of a dataset is then the \textit{empirical} average those features. In this work we focus on the random Fourier features sketch where $\Phi(\Vec{x}) := e^{\im \Omega^T\Vec{x}}$, \ie the sketch of $X$ is
\begin{equation}
    \ts \Vec{z} := \cl A\left(\frac{1}{n}\sum_{i=1}^n \delta_{\Vec{x}_i}\right) =  \frac{1}{n} \sum_{i=1}^n e^{\im \Omega^T \Vec{x}_i} \in \bb C^m,
\end{equation}
where $\delta_{\Vec{x}}$ is the Dirac measure at $\Vec{x}$ and $\Omega = (\Vec{\omega}_j \in \bb R^d)_{j=1}^m$ is a set of $m$ ``frequencies'' generated i.i.d. according to some law. To avoid over-sampling large frequencies (due to the curse of dimensionality), a good heuristic~\cite{keriven2016GMMestimation} is to set $\Vec{\omega} = R \Vec{\varphi}$, where $\Vec{\varphi} \sim \cl U(S^{d-1})$ is a normed random direction and $R \sim p_R(R;\sigma)$ is the norm of $\Vec{\omega}$. In~\cite{keriven2016GMMestimation}, this latter distribution is either (FG) a folded Gaussian $p_R \propto e^{-(\sigma R)^2}$, or (AR) the adapted radius distribution defined as $p_R \propto \left((\sigma R)^2 + \frac{(\sigma R)^4}{4}\right)^{\frac{1}{2}}e^{-(\sigma R)^2}$; in both cases $\sigma$ is a scale parameter, which should be adjusted to the current dataset either by prior knowledge or from a fast heuristic (see Sec.~\ref{sec:sigma}).

% Decoding in practice and where it could go wrong
Learning from the sketch is formulated as an inverse problem, where the ``signal'' to recover is a density $\cl P_{\Vec{\theta}}$; for example, in Gaussian mixture modeling $\cl P_{\Vec{\theta}} = \sum_{k=1}^K \alpha_k \cl N(\Vec{\mu}_k, \Sigma_k)$ with $\Vec{\theta} = (\alpha_k,\Vec{\mu}_k,\Sigma_k)_k$ the GMM parameters~\cite{keriven2016GMMestimation}, and in k-means clustering $\cl P_{\Vec{\theta}} = \sum_{k=1}^K \alpha_k \cl \delta_{\Vec{c}_k}$ with $\Vec{\theta} = (\alpha_k,\Vec{c}_k)_k$ the set of (weighted) centroids $\Vec{c}_k$~\cite{keriven2016compressive}.
The model $\wh{\Vec{\theta}}$ estimated from $\Vec{z}$ is then found using the ``sketch matching'' cost function $\cl L$,
\begin{equation}
    \wh{\Vec{\theta}} \in \arg \min_{\Vec{\theta}} \cl L(\Vec{\theta};\Vec{z})  \text{ where } \cl L(\Vec{\theta};\Vec{z}) := \| \Vec{z} - \cl A(\cl P_{\Vec{\theta}}) \|_2^2.
\end{equation}

\noindent The cost $\cl L(\Vec{\theta};\Vec{z})$ is non-convex, hence challenging to optimize exactly. In practice, a heuristic algorithm is used to solve it approximately, \ie a decoder $\Delta : \Vec{z} \mapsto \Vec{\theta}_{\Delta} \simeq \arg\min_{\Vec{\theta}} \cl L(\Vec{\theta};\Vec{z})$. For k-means and GMM, the standard decoder is \texttt{CLOMPR} (CL by Orthogonal Matching Pursuit with Replacement). It starts from an empty solution and greedily adds new atoms $\Vec{\theta}_k'$ (\ie a single Gaussian or centroid) to the current solution $\Vec{\theta}$, where new atoms $\Vec{\theta}_k'$ are found by maximizing the non-convex cost $\langle \cl A(\cl P_{\Vec{\theta}_k'}), \Vec{z}-\cl A(\cl P_{\Vec{\theta}}) \rangle$ starting from a random intial point. To increase the chances of success, $T$ trials of \texttt{CLOMPR} can be run independently and the solution with the lowest cost is selected, an approach we call \texttt{CLOMPRxT}. This decoder showed convincing empirical success for both GMM~\cite{keriven2016GMMestimation} and k-means~\cite{keriven2016compressive}. For k-means specifically, the \texttt{CLAMP} decoder (Compressive Learning with Approximate Message Parsing) was also been proposed \cite{byrne2019sketched}, which allowed to recover the centroids with lower sketch sizes; in this work we focus mainly on \texttt{CLOMPR}.

% Problem statement
Previous works thus observed that compressive learning (\eg using \texttt{CLOMPR}) performed well under the right circumstances (\eg when the sketch size $m$ is large enough, and its frequency sampling pattern $\Lambda$ is well-chosen). \textit{But is it possible to improve the existing CL schemes further? And if so, how? To perform constructive research on these question, we must first understand when and why existing compressive learning schemes fail.} In this work, we provide some insights on this question, based on observations from numerical simulations.
\vspace{-0.5em}

\section{Results}
\paragraph{CL failure scenarii}

\begin{figure}
\centering
    \includegraphics[width=0.77\linewidth]{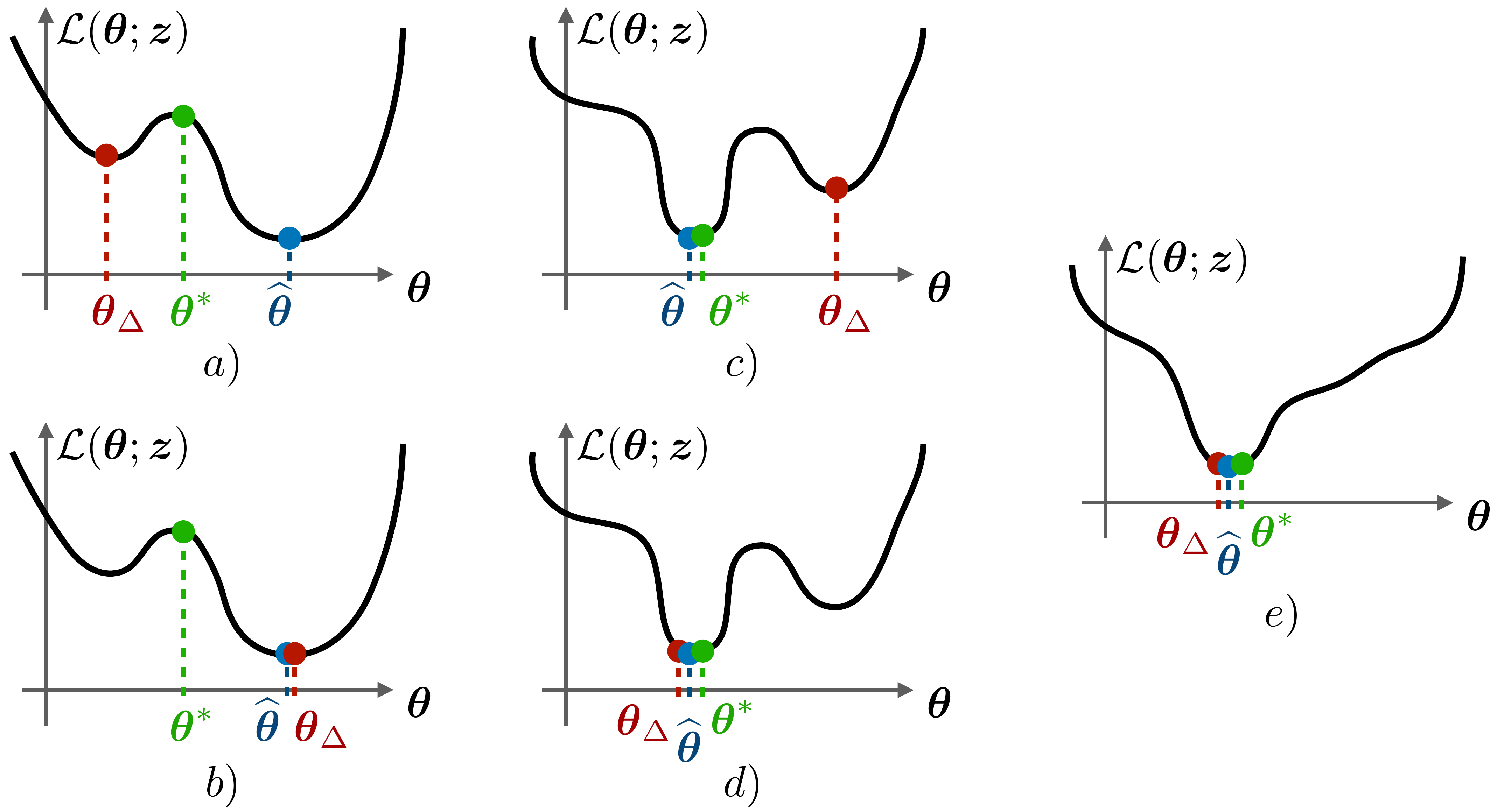}
	\caption{Five scenarii in compressive learning: the cost function $\cl L$ (determined by the sketch $\Vec{z}$) can be bad (when $\wh{\Vec{\theta}} \neq \Vec{\theta}^*$, cases $(a)$ and $(b)$), acceptable (when $\wh{\Vec{\theta}} \simeq \Vec{\theta}^*$, but with a small basin of attraction, cases $(c)$ and $(d)$), or good ($\wh{\Vec{\theta}} \simeq \Vec{\theta}^*$ with a large basin of attraction, case $(e)$). Given this cost function, the decoder $\Delta$ can either fail to find the global optimum (when $\Vec{\theta}_{\Delta} \neq \wh{\Vec{\theta}}$, cases $(a)$ and $(c)$) or ``succeed'' (when $\Vec{\theta}_{\Delta} \simeq \wh{\Vec{\theta}}$, the latter being possibly different from $\Vec{\theta}^*$, cases $(b)$, $(d)$ and $(e)$).}
	\label{fig:costs}
\end{figure}

In Fig.~\ref{fig:costs}, we classify the different outcomes $(a)-(e)$ that are possible when one runs a compressive learning decoder on a given sketch, sorted by ``what could go wrong''. If the sketch was poorly designed, then $\cl L$ does not indicate (through its global minimum $\wh{\Vec{\theta}}$) the desired parameters $\Vec{\theta}^*$, as shown in cases $(a)-(b)$. In~\cite{gribonval2017compressiveStatisticalLearning}, the authors derive theoretical guarantees on (an upper bound for) the probability that this failure occurs, as a function of the sketch size $m$ and the compatibility between the sketching function and the target learning task. However, even if the sketch matching cost function aligns with the ideal parameters, it is possible that the decoder fails to solve the non-convex problem, as shown in case $(c)$ (the existing decoders do not have any guarantee). Whether or not the decoder succeeds depends on the general shape of the cost $\cl L$ (besides the position of its global optimum). In \texttt{CLOMPR} for example, convergence to $\wh{\Vec{\theta}}$ is ensured if the random initialization point falls into its \textit{basin of attraction}. It was shown (for the centroid selection in k-means) that the size of this basin of attraction increases with the sketch size $m$~\cite{traonmilin2019basins}. To understand how to improve current CL schemes, we would like to classify practical failures of decoders into either scenarii $(a)-(b)$ (where we know that performance improvement is possible---and, for $(b)$, necessary---by changing the sketch function) or scenario $(c)$ (where we know that improvement is possible by changing the decoder). In the follow, we analyze how this classification is modified when two parameters of the sketch are affected, namely the sketch size $m$ (Sec.~\ref{sec:size}) and scale parameter $\sigma$ in the frequency sampling (Sec.~\ref{sec:sigma}).

\vspace{-0.5em}
\subsection{Influence of the sketch size}
\label{sec:size}
\begin{figure}
\centering
    \includegraphics[width=0.71\linewidth]{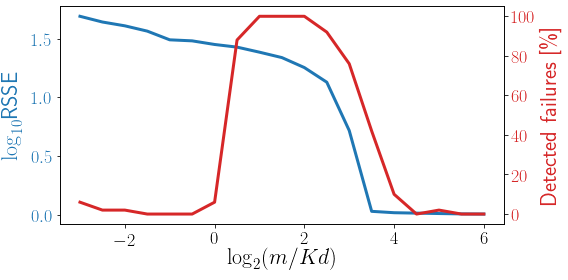}
	\caption{\textbf{Blue}: relative SSE (with respect to \texttt{k-means}) of the compressively learned centroids $\Vec{\theta}_{\texttt{CLOMPRx10}}$, as a function of the sketch size $m$ (log scale, median over 50 draws of $\Omega$). \textbf{Red}: number of times a failure of the type $\cl L(\Vec{\theta}_{\texttt{CLOMPRx10}};\Vec{z}) > \cl L(\Vec{\theta}^*;\Vec{z})$ was detected. The data is $n = 5 \times 10^4$ points drawn from $K = 10$ Gaussians in $d = 10$.}
	\label{fig:detectFailures}
\end{figure}

In this section we focus on the k-means problem, where the goal is to position $K$ centroids $\Vec{c}_k$ such that the Sum of Squared Errors, $\mathrm{SSE}(\{\Vec{c}_k\}_{k=1}^K) := \sum_i \min_k \| \Vec{x}_i - \Vec{c}_k \|_2^2$, is minimized. In Fig.~\ref{fig:detectFailures} we plot the quality of compressively learned centroids (by \texttt{CLOMPR} with 10 trials) as a function of $m$; as expected from previous works, they are always perfectly recovered when $m \geq 10Kd$, which corresponds to case $(e)$. We also compared the cost of those centroids $\cl L(\Vec{\theta}_{\texttt{CLOMPRx10}})$ with the cost of the ground-truth centroids $\cl L(\Vec{\theta}^*)$, and reported the average number of times that $\cl L(\Vec{\theta}_{\texttt{CLOMPRx10}}) > \cl L(\Vec{\theta}^*)$. This allows us to identify an additional regime: when $m \leq Kd$, we know that we are in case $(a-b)$, \ie the cost is ill-defined (note that here the condition $m \leq Kd$ coincides with the over-parameterized regime where there are less measurements than degrees of freedom). In between ($1 \leq m/Kd \leq 10$), the situation is less clear: as the number of measurements increase, we transition from $(a-b)$ to $(c)$ with gradually more $(d)$, then finally to $(e)$, but it's not clear at which point the $(a)-(b)$ switches to $(c)$. 

\begin{figure}
\centering
    \includegraphics[width=0.85\linewidth]{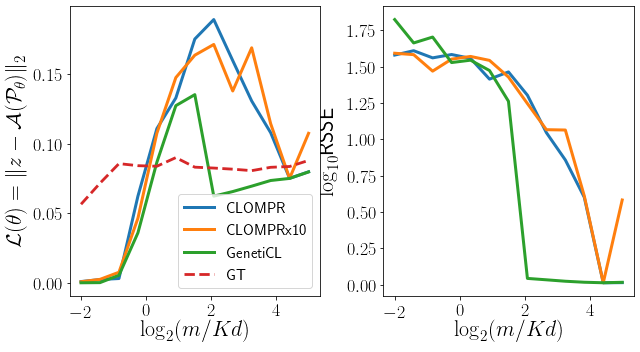}
	\caption{\textbf{Left}: cost function $\cl L(\Vec{\theta};\Vec{z})$ as a function of an increasing sketch size $m$ (for a fixed draw of $\Omega$), evaluated on the centroids $\Vec{\theta}_{\Delta}$ obtained by different decoders ($\Delta \in \{ \texttt{CLOMPR}, \texttt{CLOMPRx10}, \texttt{GenetiCL}\}$) as well as the on ``ground-truth'' centroids $\Vec{\theta}^*$. \textbf{Right}: relative SSE corresponding to those centroids $\Vec{\theta}_{\Delta}$. Data are $n = 10^5$ samples from $K = 10$ Gaussians in $d = 5$.}
	\label{fig:gen}
\end{figure}

To investigate this transition further, we perform another experiment, where this time we keep $\Omega$ fixed to analyze only the performances of the decoder for a fixed cost: this is shown Fig.~\ref{fig:gen}. Moreover, we would like to approach the global minimizer $\wh{\Vec{\theta}}$ as much as possible. For this purpose, we propose a new genetic-algorithm-based decoder, that we name \texttt{GenetiCL}. Its principle is to maintain a population of ``chromosome'' candidate solutions $\Vec{\theta}$ that explore the optimization landscape by random ``mutations'' (noise) and combinations by ``crossover'' (swapping centroids) based on a fitness score that we set to be $\|\Vec{z} - \cl A(\cl P_{\Vec{\theta}})\|_2^{-\gamma}$ for $\gamma > 0$.
While orders of magnitude slower than \texttt{CLOMPR} (we discourage using it for anything other than research purposes), it finds more promising (local) minimizers to $\cl L$, see Fig.~\ref{fig:gen}. For $m \geq 2Kd$, \texttt{GenetiCL} performs better, and we are for sure in case $(c)$ for \texttt{CLOMPR}. However, uncertainty remains in the zone $Kd \leq m \leq 2Kd$, as the ground truth cost is lower than what the decoders can find, hence it's not certain whether we are in case $(c)$ or $(a)$.

\vspace{-0.5em}

\subsection{Influence of the scale and task}
\label{sec:sigma}
%Careful: we neglect our findings from the previous section.

\begin{figure}
\centering
    \includegraphics[width=0.85\linewidth]{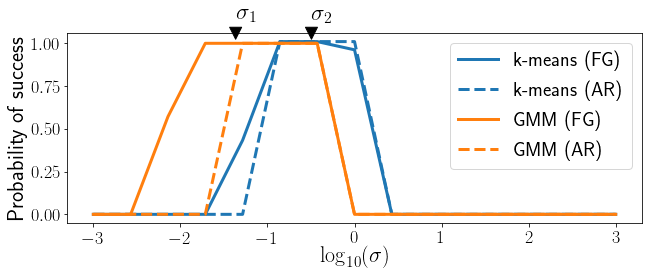}
	\caption{Empirical success rate of \texttt{CLOMPRx3} when solving k-means (blue, defining success as $\mathrm{RSSE} \leq 1.3$) and GMM fitting (orange, defining success as $\cl P_{\Vec{\theta}}(X) \geq (1.3)^{-1} \cl P_{\Vec{\theta}^*}(X)$ with $\cl P(X)$ the likelihood of $X$ given $\cl P$) from a sketch with $m = 20Kd$ frequencies drawn from the FG (plain) and AR (dashed) distributions with varying scale $\sigma$. The heuristic proposed in~\cite{keriven2016GMMestimation} for GMM (resp. in~\cite{byrne2019sketched} for k-means) yields $\sigma_1$ (resp. $\sigma_2$), indicated by black triangles. Data are $n = 10^5$ samples from $K = 6$ Gaussians in $d = 5$.}
	\label{fig:sigma}
\end{figure}
We now analyse how the cost function $\cl L$ changes with the sketch scale $\sigma$, as well as with the considered task (solving k-means or GMM from a same sketch), shown Fig.~\ref{fig:sigma}. First, notice that there exist scales ($\sigma \in [1, 2]$ in our case) where k-means succeeds but GMM fails. Intuitively, the culprit is the sketch (case $(b)$), which captures a very coarse view of the data distribution; it cannot estimate the clusters shape precisely, but can still locate them. Conversely, there are also scales ($\sigma \in [-3, -1.5]$ in our case) where GMM succeeds but k-means fails.
This is more surprising, as the successfully extracted GMM contains the centroids (in the means of the Gaussians); we thus expect that the decoder failed (case $(c)$), which would mean that switching the task from k-means to GMM leads to a better cost function (going from $(c)$ to $(e)$).
Plotting the detected failures (as in Fig.~\ref{fig:detectFailures}, not shown here for conciseness) supports these explanations: when $\sigma \in [-3, -1.5]$ we detect failures of the k-means decoder (meaning that $(c)$ is possible), but when $\sigma \in [1, 2]$ not such failures are detected from the GMM decoder (meaning we are for sure in $(a)-(b)$).

%We conclude with a few words on choosing the frequency sampling heuristic in practice: neither of the heuristics  proposed in previous works (for GMM and KM independently), hit the spot where \textit{both} k-means and GMM succeed; we suggest using the geometrical mean of both heuristics when no a priori knowledge of the task is given. On a side note, we also remark that the AR distribution is more sensitive to the choice of $\sigma$, and recommend using FG when $\sigma$ is coarsely estimated.

%This was predicted by the theory but not shown in practice.

%\subsection{Pre-sketch-based VS a priori heuristics}
%Sounds good, doesn't work

%\section{Conclusion}
%\label{sec:conclusion}
% 0.25 column

%\paragraph{Moving forward}
% using theory
% using other strategies

%% You can make the bibliography smaller
\printbibliography

\end{document}